\documentclass[conference]{IEEEtran}
\IEEEoverridecommandlockouts
\usepackage{cite}
\usepackage{amsmath,amssymb,amsfonts}
\usepackage{algorithmic}
\usepackage{graphicx}
\usepackage{textcomp}
\usepackage{xcolor}
\usepackage{hyperref}

\def\BibTeX{{\rm B\kern-.05em{\sc i\kern-.025em b}\kern-.08em
    T\kern-.1667em\lower.7ex\hbox{E}\kern-.125emX}}
\begin{document}

%\title{Memory-Efficient Graph Convolutional Networks for Event Processing}
\title{Memory-Efficient Graph Convolutional Networks for Object Classification and Detection with Event Cameras
\thanks{The work presented in this paper was supported by the programme ``Excellence Initiative -- Research University'' for the AGH University of Krakow and partly by Sorbonne University}
}

%\author{\IEEEauthorblockN{Kamil Jeziorek}
%\IEEEauthorblockA{\textit{Embedded Vision Systems Group,} \\
%\textit{Computer Vision Laboratory,} \\
%\textit{Department of Automatic Control and Robotics,} \\
%\textit{AGH University of Krakow, Poland}\\
%\textit{\href{mailto:kjeziorek@student.agh.edu.pl}{kjeziorek@student.agh.edu.pl}}}
%\and
%\IEEEauthorblockN{Tomasz Kryjak, Senior Member, IEEE}
%\IEEEauthorblockA{\textit{Embedded Vision Systems Group,} \\
%\textit{Computer Vision Laboratory,} \\
%\textit{Department of Automatic Control and Robotics,} \\
%\textit{AGH University of Krakow, Poland}\\
%\textit{\href{mailto:tomasz.kryjak@agh.edu.pl}{tomasz.kryjak@agh.edu.pl}}}
%}

\author{
\IEEEauthorblockN{Kamil Jeziorek$^*$, Andrea Pinna $^\dag$, Tomasz Kryjak $^*$$^\dag$}
\IEEEauthorblockA{$^*$\textit{Embedded Vision Systems Group,} 
\textit{Department of Automatic Control and Robotics,} 
\textit{AGH University of Krakow, Poland}}
\IEEEauthorblockA{$^\dag$\textit{Sorbonne Universite, CNRS, LIP6, F-75005 Paris, France}}
\textit{\href{mailto:kjeziorek@student.agh.edu.pl}{kjeziorek@student.agh.edu.pl}, \href{mailto:andrea.pinna@lip6.fr}{andrea.pinna@lip6.fr}, \href{mailto:tomasz.kryjak@agh.edu.pl}{tomasz.kryjak@agh.edu.pl}}
}

\maketitle
%%%%%%%%%%%%%%%%%%%%%%%%%%%%%%%%%%%%%%%%%%%%%%%%%%%%%%%%%%%%%%%%%%%%%%%%%%%%%%%%%
W\begin{abstract}
Recent advances in event camera research emphasize processing data in its original sparse form, which allows the use of its unique features such as high temporal resolution, high dynamic range, low latency, and resistance to image blur. One promising approach for analyzing event data is through graph convolutional networks (GCNs). However, current research in this domain primarily focuses on optimizing computational costs, neglecting the associated memory costs.
In this paper, we consider both factors together in order to achieve satisfying results and relatively low model complexity. For this purpose, we performed a comparative analysis of different graph convolution operations, considering factors such as execution time, the number of trainable model parameters, data format requirements, and training outcomes. Our results show a 450-fold reduction in the number of parameters for the feature extraction module and a 4.5-fold reduction in the size of the data representation while maintaining a classification accuracy of 52.3\%, which is 6.3\% higher compared to the operation used in state-of-the-art approaches.
To further evaluate performance, we implemented the object detection architecture and evaluated its performance on the N-Caltech101 dataset. The results showed an accuracy of 53.7\% mAP@0.5 and reached an execution rate of 82 graphs per second.
\end{abstract}

%%%%%%%%%%%%%%%%%%%%%%%%%%%%%%%%%%%%%%%%%%%%%%%%%%%%%%%%%%%%%%%%%%%%%%%%%%%%%%%%%
\section{Introduction}\label{introduction}

Event cameras are modern vision sensors whose operating principle is inspired by the human eye.
Unlike traditional cameras, which record frames at fixed intervals, event cameras detect changes in light intensity in individual pixels, resulting in the generation of an asynchronous stream of information, into so-called events. 
This unique design offers several advantages, including a high tonal range, high temporal resolution and high resistance to motion blur effects. 
As a result, event cameras have gained a lot of attention and have found applications in a variety of challenging scenarios where conventional video cameras face limitations, particularly in advanced mobile robotics (autonomous vehicles: drones, cars) and for broadly improving video capture (improvement in uneven lighting conditions, increasing frame rates) \cite{PAMI}. 

Despite their clear advantages, event camera data processing is difficult due to its sparse and spatial-temporal nature, which requires approaches that differ from those developed for traditional vision systems. 
Early approaches involved projecting events into dense two-dimensional pseudo-representations, such as event frames, or reconstructing them into greyscale frames using deep neural networks. 
Although these methods have made it possible to exploit the potential of convolutional neural networks, they have drawbacks. 
First of all, the projection or reconstruction process loses the key features of event cameras, specifically the high temporal resolution and the advantages of data sparsity (computational and energy efficiency).
In addition, the reconstruction process is quite computationally complex, as shown in the work \cite{reconstruction_e2vid}.

This has prompted researchers to look for alternative solutions for processing event data while keeping it in a sparse form.
The first attempts to solve this problem involved filter-based methods \cite{filters} and spiking neural networks (SNNs) \cite{snn}. However, filter-based methods require manual definition of equations, which makes it difficult to achieve good results for more complex tasks. SNN-based methods, on the other hand, still have underdeveloped learning rules, and their implementation can be more complicated compared to traditional convolutional networks.
Another relatively new proposal is the use of graph neural networks.
Recent advances in this field have shown that event processing using graph convolutional networks is possible \cite{b1} \cite{b2} \cite{b3}. The undoubted advantage of this approach is to process event data in the form of the original point cloud while exploiting the potential of convolution operations.
However, we saw that existing approaches focused primarily on reducing computational complexity, placing less importance on memory efficiency. 

% W niniejszym artykule postawiliśmy sobie za cel przeanalizowanie wpływu wykorzystania sieci grafowych na złożoność pamięciową. W tym celu porównaliśmy różne operacje konwolucji grafowych pod względem wykorzystywanych danych do przetwarzania oraz rozmiaru sieci. Wyniki pokazały, że istnieje możliwość redukcji obu współczynników przy zachowaniu stosunkowo dobrych wyników w porównaniu do najnowszych badań. Ta praca stanowi uzupełnienie istniejącej wiedzy, ponieważ dotychczasowi autorzy skupiali się głównie na optymalizacji złożoności obliczeniowej.

In this paper, our objective was to analyse the impact of employing graph networks on memory complexity. To achieve this, we conducted a comparison of various graph convolution operations based on their data requirements and network size. The findings demonstrated the potential to decrease both data and network size while still achieving satisfactory performance in comparison to state-of-the-art results. This study serves as a valuable addition to current knowledge since previous researchers have primarily concentrated on only optimising computational complexity.

%TODO Przerobić Niniejszy artykuł przedstawia nowe spojrzenie na optymalizację pamięci w przetwarzaniu danych zdarzeń, biorąc pod uwagę zarówno wydajność, jak i złożoność obliczeniową. 

%TODO Graphical abstraact - opisać jak go widzę

The remainder of the paper is organised as follows. Section \ref{event_cameras} presents event cameras, their advantages and the mechanism of information generation. Section \ref{gnn} describes graph convolutional networks and the principle of convolution and pooling operations. Section \ref{related} discusses previous work on event processing using graph convolutional networks and their memory usage problems. Section \ref{experience} presents the experiments we conducted to demonstrate improvements in memory usage. Finally, we summarise our results and present future research plans.

%%%%%%%%%%%%%%%%%%%%%%%%%%%%%%%%%%%%%%%%%%%%%%%%%%%%%%%%%%%%%%%%%%%%%%%%%%%%%%%%%
\section{Event Cameras}\label{event_cameras}

Event cameras, also known as Dynamic Vision Sensors (DVS), are bio-inspired cameras. Their unique operating principle offers a number of advantages over traditional cameras, changing the way visual information is captured and processed.

One of the distinctive features of event cameras is their asynchronous nature. 
Unlike conventional cameras, which record brightness levels globally for all pixels at fixed intervals, event cameras record brightness changes for individual pixels. 
This independence at the pixel level enables event cameras to capture precise temporal information, as events are only generated when a significant change in the brightness of a given pixel occurs (due to movement in the scene, the cameras' own movement or a change in lighting). 
Consequently, event cameras excel at capturing dynamic scenes with minimal motion blur. 
The event generation process in DVS is controlled by a threshold mechanism. When the logarithmic change in light intensity \(\Delta L\) at a specific pixel \((x_i, y_i)\) exceeds a predefined threshold value \(C\), an event is generated. This can be expressed as:

\begin{equation}
\Delta L(x_i,y_i,t_i) = L(x_i,y_i,t_i) - L(x_i, y_i, t_i - t_i^*) \geq p_i C \label{generate}
\end{equation}

where \(t_i\) represents the time of occurrence of the current event and \(t_i^*\) the previous event. This equation defines the basic principle of event generation in event cameras, ensuring that only significant and meaningful changes are recorded. Each generated event includes four items of information: the pixel coordinates \((x_i, y_i)\), the exact timestamp \(t_i\) derived from the internal clock, and the polarity \(p_i\) indicating whether the change in light intensity was positive or negative. These events together form a spatial-temporal sequence of asynchronous data:

\begin{equation}
E = \{e_i\}_{i=1}^N, e_i = (x_i, y_i, t_i, p_i) 
\label{events}
\end{equation}

Moreover, thanks to their high dynamic range (over 120 dB), event cameras are able to faithfully reproduce scenes containing both bright and dark areas, capturing a wide range of light intensities. These features make event cameras particularly well suited to difficult lighting conditions, such as poorly lit environments or scenes with high contrast (uneven lighting).

%%%%%%%%%%%%%%%%%%%%%%%%%%%%%%%%%%%%%%%%%%%%%%%%%%%%%%%%%%%%%%%%%%%%%%%%%%%%%%%%%
\section{Graph Convolutional Networks}\label{gnn}

Graph convolutional networks (GCNs) \cite{gcns} are a type of machine learning models that operate on graph-structured data. Unlike traditional neural networks, which are mainly designed for grid data, GCNs are able to extract the complex relationships and dependencies present in graph data. A graph is made up of vertices \(V\) and edges \(\mathcal{E}\), which represent the relationships between two vertices. GCN uses this graph structure to learn and propagate information between nodes, enabling them to predict or perform tasks based on existing connection patterns.

\subsection{Graph Convolution}\label{conv}

The convolution operation is a key element of graph neural networks. It allows the propagation of information between vertices in order to update their representation and obtain key information about the graph. The general principle of this operation is referred to as neighbourhood \(N(i)\) aggregation and is performed for each vertex \(x_i\) in three steps: message, aggregation and update function.

The message function is responsible for generating the information passed between vertices. It is defined at single edge level between vertices \(x_i\) and \(x_j\) and can take into account edge attributes \( e_{i,j}\) if they are defined:

\begin{equation}
msg_{i,j} = \phi (x_i, x_j, e_{i,j}) \label{msg}
\end{equation}

The aggregation function collects forwarded messages from neighbouring vertices \(x_j\), which were obtained by the message function, and combines them at the level of a single vertex \(x_i\):

\begin{equation}
agr_i = \bigoplus_{j \in N(i)} msg_{i, j} \label{agr}
\end{equation}

Finally, the update function is executed, which is responsible for updating the vertex representation \(x_i\). In this phase, the vertex representation is updated based on the aggregated information from its neighbours.

\begin{equation}
x_i' = \gamma (agr_i) \label{update}
\end{equation}

where the functions \(\phi\) and \(\gamma\) can be represented as linear or MLP functions, while the operation \(\bigoplus\) can be specified as a sum, mean or maximum value function. It is worth noting that message, aggregation and update functions can be defined in different ways depending on the specific model of graph convolutional networks. Different GCN variants may use different operations and transformations in these functions, adjusted to the specific requirements and structure of the graph data. The impact of different convolution variants for event data processing is the topic of this article. 

\subsection{Graph Pooling}\label{pool}

Pooling layers in GCNs serve a similar purpose to their equivalents in convolutional neural networks: reducing the dimensionality of the data while retaining key information and generalisation.

In GCNs, the pooling layer involves grouping vertices based on a regular grid of a specific dimension \cite{voxel_grid} or using methods to cluster points in space \cite{cluster}. 
The pooling process involves selecting representative vertices or aggregating information within each group or cluster. This selection can be based on different criteria, such as the importance of a vertex, the maximum value or the average among all vertices. The result of this operation is a new vertex that represents the group.
By creating denser representations through pooling, GCN reduces the computational complexity in subsequent layers. This allows for more efficient processing of large-scale data, while preserving essential features and information flow. Furthermore, it also allows the use of fully connected layers for data classification.

%TODO: Przydałby się rysunek do tego... takich schemacik ilustrujący grafy, konwolucje i pooling

\section{Previous works}\label{related}

This review focuses exclusively on graph convolutional networks applied to event data processing. Due to the relatively new approach, the amount of work on this topic is still small. 

The initial works \cite{standard1} \cite{standard2} focused on constructing a graph from incoming events and processing it in its entirety using graph convolutional networks. 
Although these operations are only performed on the relevant event information, which reduces the computational costs associated with dense zero-valued representations (as described in the \ref{event_cameras} section), they require repeated operations on previous events that are not affected by new events.
In practice, this means that it is necessary to process the entire graph from the beginning for new events, even if most of the information is unchanged. Such computational redundancy can lead to inefficient use of computing resources and time constraints.

Recent researches have focused on applying graph neural networks to events, updating the network asynchronously event by event. The aim of such an operation is to perform computations only on changing events, thus increasing computational efficiency. In a series of publications \cite{b1}, \cite{b2}, \cite{b3}, a technique has been proposed in which propagation is performed for a single event. A new event is merged with an existing graph and the neighbouring vertices are updated. The remaining vertices that do not have a new neighbour do not need to be updated, leading to a reduction in the number of operations.

However, in order to perform this operation, the layers in these networks require information about their current state. 
In the example described in the paper \cite{b1}, a particular subset of events is selected from the entire set of events, which is then passed through the network as a simple graph to initialise the internal layers.
Then, for each new incoming event, the neighbouring vertices that need to be updated are identified, and the result of this operation is propagated to the next layers. 
As each layer has its own copy of the graph, the memory required to store it is considerable. 
The memory requirement increases with the number of layers, the size of the camera images (there are currently cameras with a resolution of 1 megapixel, i.e. HD 1280 x 720 pixels \cite{b4}) and the number of vertices per graph.
The total memory required to store all copies may exceed the capacity of embedded systems.

On the other hand, events in their original form consist of four values (\(x, y, t, p\)), as described in \ref{event_cameras}. When the graph is created, each event is represented as a vertex with a 3D spatio-temporal location (\(x, y, t \)) and an attribute \(p\), while edge definition requires the indexes of two vertices to be determined.
As mentioned in the \ref{conv} section, there are different versions of the graph convolution operation, some of which only consider vertex attribute information, and edge indices, while others also consider edge-specific attributes. 
In the publications \cite{b1}, \cite{b2}, \cite{b3} applied convolution function requires these values to be taken into account.
This is achieved using a Cartesian coordinates of linked nodes in which the resulting value contains additional information. 
As the number of edges increases, the number of their attributes also increases, leading to a larger memory requirement for data storage and processing.

While all work focuses on minimising computational costs, our aim is to present another aspect that involves the usage of memory resources. 
This includes both the data generated during graph construction and the network architecture, while also aiming for good performance and low processing times.
%%%%%%%%%%%%%%%%%%%%%%%%%%%%%%%%%%%%%%%%%%%%%%%%%%%%%%%%%%%%%%%%%%%%%%%%%%%%%%%%%

\section{Experiments}\label{experience}

% Wszystkie przedstawione eksperymenty zostały przeprowadzone na zbiorze danych N-Caltech101 \cite{b5}, który składa się ze 100 klas (oryginalna wersja ma 101 klas, w tym "Faces" i "Faces Easy", które zostały połączone w wersji neuromorficznej). 
% Obejmuje on 8246 próbek, przy czym wymiary danych nie przekraczają 240 x 180. 
% Dane te pozwalają na zastosowanie ich zarówno w zadaniach klasyfikacji, jak i detekcji, ponieważ każda próbka zawiera pojedynczy obiekt. %TODO ja bym użył klasyfikacji i detekcji - ok
% Podczas procesu uczenia dane zostały losowo podzielone na 80\% zbioru treningowego i 20\% zbioru testowego, zapewniając równy rozkład próbek z każdej klasy.

% Do przeprowadzenia eksperymentów wykorzystaliśmy bibliotekę PyTorch Geometric \cite{b6}, która zapewnia wygodne narzędzia i funkcje do pracy z danymi o strukturze grafu i umożliwia płynną integrację z architekturami grafowych sieci konwolucyjnych.

% All the presented experiments were conducted on the N-Caltech101 \cite{b5} dataset, which consists of 100 classes (the original version has 101 classes, including 'Faces' and 'Faces Easy', which were combined in the neuromorphic version). 

All the presented experiments were conducted on the N-Caltech101 dataset \cite{b5}, which was created by using a~moving event camera to capture the standard Caltech101 dataset displayed on a monitor. The N-Caltech101 dataset consists of 100 classes, where the original version of the dataset includes 101 classes, including 'Faces' and 'Faces Easy', which were merged in the neuromorphic version.
It includes 8246 samples, with data dimensions not exceeding 240 x 180. 
This data allows it to be used in both classification and detection tasks, as each sample contains a single object.
During the training process, the data was randomly split into an 80\% training set and a 20\% test set, ensuring an equal distribution of samples from each class.

For the experiments, we used the PyTorch Geometric library \cite{b6}, which provides convenient tools and features for working with graph-structured data and allows seamless integration into graph convolutional network architectures.

\subsection{Graph size}\label{graph_size}

\begin{figure}[!t]
\centerline{\includegraphics[scale=0.37]{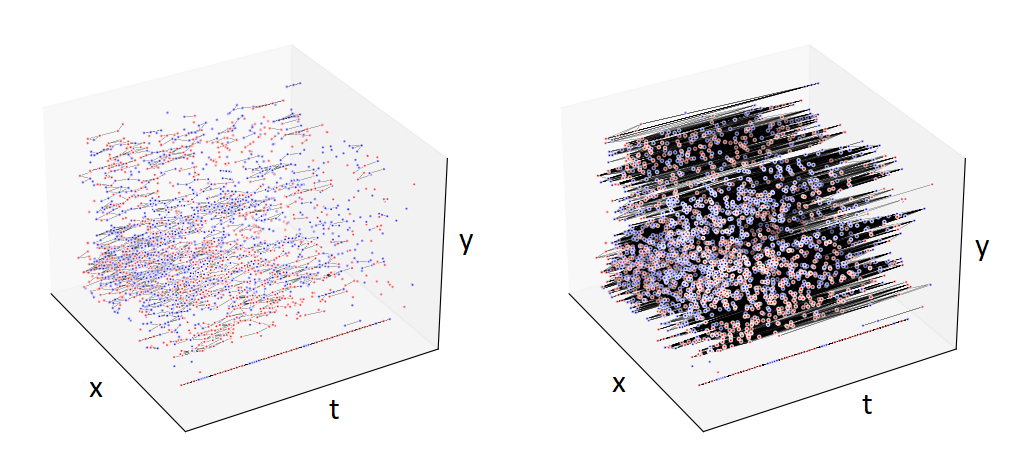}}
\caption{Presentation of the generated edges of the graph: On the left is the data after normalisation, where edges only connect to the nearest vertices. On the right is the graph without normalisation, where there are more edges, spanning along the entire \textit{time} axis.} 
\label{fig}
\end{figure}

To assess the size of the data required by the graph, we conducted an experiment on the entire dataset. We used the method outlined in \cite{b2} to create a graph for each sample. We set the maximum number of events per graph equal to 25 000, where the events with the highest density within a certain time window were selected. We set the neighbourhood radius relative to which vertices were connected by edges to 5, and to limit the number of edges generated, we set the neighbour limit to 32 per vertex. We tested two different approaches: in the first, we normalised the occurrence times of the events to values close to the resolution of the data (in the range of 0 to 100), and in the second we did not normalise, where the time values were in the range of microseconds. Normalising the time has a significant impact on the training results, but also affects the size of the data. If the time values are increased, the vertices are further apart, resulting in local linking of events based on a defined neighbourhood radius and a reduction in the number of generated edges. 
Figure \ref{fig} illustrates the differences between the two approaches.

For the entire N-Caltech101 dataset, the average number of events is 24,457 (the lower value is due to samples with event counts not exceeding 25,000 events), resulting in 24,457 vertex \(p\) and spatial-temporal (\(x, y, t\)) values. The average number of edges for the unnormalised data is 756,744, while for the normalised data it is 381,563.

The PyTorch Geometric library represents vertex values as \textit{int32} and spatial-temporal positions as \textit{float32} vectors. Each edge is represented as a two-element vector with the vertex indices as \textit{int32} values. Finally, the edge attribute value consists of three \textit{float32} values. 
In total, each vertex requires 16 bytes, while each edge, including the attribute, requires 40 bytes. 
The average amount of memory used by the graph is about 15 MB for normalised data, compared to 30.5 MB for unnormalised data. 

If the edge attribute values in the graph are excluded, the data size for normalised graphs decreases to 3.5 MB, while for unnormalised data it is 6.5 MB. Thus, omitting the edge attribute values reduces the data size by an average of more than 4.5 times. In comparison, a 240 by 180 pixel raw image with a single channel using the \textit{uint8} data type takes up 0.04 MB, while a 3 channel image takes up 0.12 MB. This shows how much memory such data representations require and how important it is to reduce them. 

It is important to note that the above values are specified for low resolution and a limited number of events in the graph. For larger data sizes generated with, for example, a 1 megapixel camera \cite{b4}, the number of events in the graph, and thus the number of edges, will increase significantly. To remedy this, the size of the input data can be rescaled, as is often practised with classical convolutional networks (CNNs). 

In this experiment, we only focused on omitting edge attributes to reduce memory. However, it is also possible to reduce memory by using smaller data types than those used in the PyTorch Geometric library. For example, for a graph with a maximum number of events of 25,000, vertex index values can be represented as 15-bit numbers in FPGA, which may be a topic for future research.

\subsection{Convolution operations comparison}\label{classification}

In order to study the effect of skipping edge features on training results, we conducted a comparison of several convolution operations.
As a reference, we chose the SplineConv convolution operation \cite{spline}, which is used in recent publications  \cite{b2} \cite{b3}. It uses both vertex and edge features. 
We chose two standard operations as models that do not consider edge features: GCNConv \cite{gcn} and SAGEConv \cite{sage}, which only require edge indices and vertex features, and two convolutional operations whose main purpose is to process point cloud data -- EdgeConv \cite{edge} and PointNetConv \cite{pointnet}.

The EdgeConv function uses only information about vertex features and edge indices, and the convolution operation is defined as:

\begin{equation}
x_i' = \sum_{j\in N(i)} \phi (x_i || x_j - x_i) \label{edgeconv}
\end{equation}

The PointNetConv operator uses vertex features, edge indices but also vertex positions in space:

\begin{equation}
x_i' = \gamma(\underset{j \in N(i)}{max} \phi (x_j, p_j - p_i)) \label{pointconv}
\end{equation}

It is worth noting that PointNetConv uses the positions of vertices \(p_i\) and \(p_j\) in its message function, explained in Section \ref{conv}, to determine the distance between them, without having to first calculate the value of the edge attribute based on the Cartesian metric and storing it. 

The PyTorch Geometric library allows to define a custom function \(\phi\) for the EdgeConv model, and functions \(\phi\) and \(\gamma\) (optional) for the PointNetConv model. In our case, we decided to define single linear layers with the minimum required size. We did not use the \(\gamma\) function for the PointNetConv model.

We compared the models in terms of three key aspects: number of trainable parameters, execution time  and achieved accuracy. 
We used an identical architecture for each model, as proposed in \cite{b2}. 
It consisted of seven convolution layers with following output channels (8, 16, 32, 32, 32, 128, 128), and the fifth convolution layer was followed by a MaxPooling layer with a window size of (16, 12). The output of the feature extractor was flattened to 2048, and the classifier consisted of a single linear layer with an output size of 100, corresponding to the number of classes.

In the first step, we checked the number of parameters for each feature extraction layer and for the fully connected layer -- the results are presented in the table \ref{parameters}. 
For the SplineConv, the number of parameters needed for the feature extractor is about 20.2 million, while for the other models the values are about 41k for the GCNConv and SAGEConv models, 43k for the PointNetConv model, and 81k for the EdgeConv model. Since the number of outputs from the feature extractor is the same for all models, the fully connected layer has an identical number of parameters for every model.
This means that the reduction in the number of parameters ranges from 249 to 492 times compared to the SplineConv model.

\begin{table}[!t]
\caption{The number of learning parameters in the feature extraction layer and the fully connected layer of the models.}
\resizebox{\columnwidth}{!}{%
\begin{tabular}{|l|c|c|c|}
\hline
Layer & Feature Extraction Layers & Fully Connected Layers & Sum \\ \hline
SplineConv & 20.2 M & 204 K & 20.4 M \\ \hline
EdgeConv & 81 K & 204 K & 285 K \\ \hline
GCNConv & 41 K & 204 K & 245 K \\ \hline
SAGEConv & 41 K & 204 K & 245 K \\ \hline
PointNetConv & 43 K & 204 K & 247 K \\ \hline
\end{tabular}%
}
\label{parameters}
\end{table}

% \begin{table}[!t]
% \caption{The number of learning parameters for each layer in the model and for each convolution operation. For Spline convolution operations, the number of parameters reaches millions, while for other convolutions the number of parameters is in the tens of thousands.}
% \resizebox{\columnwidth}{!}{%
% \begin{tabular}{|l|c|c|c|c|c|c|c|c|c|}
% \hline
% Layer & Conv1 & Conv2 & Conv3 & Conv4 & Conv5 & Conv6 & Conv7 & MLP & Sum \\ \hline
% SplineConv & 8.2 K & 262 K & 525 K & 525 K & 2.1 M & 8.4 M & 8.4 M & 204 K & 20.4 M \\ \hline
% EdgeConv & 80 & 1.1 K & 2.1 K & 2.1 K & 8.6 K & 33.2 K & 33.2 K & 204 K & 285 K \\ \hline
% GCNConv & 64 & 608 & 1.1 K & 1.1 K & 4.5 K & 16.8 K & 16.8 K & 204 K & 245 K \\ \hline
% SAGEConv & 64 & 608 & 1.1 K & 1.1 K & 4.5 K & 16.8 K & 16.8 K & 204 K & 245 K \\ \hline
% PointNetConv & 112 & 704 & 1.2 K & 1.2 K & 4.9 K & 17.2 K & 17.2 K & 204 K & 247 K \\ \hline
% \end{tabular}%
% }
% \label{parameters}
% \end{table}

We then measured the execution time.  We used an NVIDIA GeForce RTX 3060 graphics card for the computation. 
We measured the time for the network processing operation alone, and the average value over the entire validation set is shown in the table \ref{times}. 
The best results were achieved by the SAGEConv model, with an average processing time of 6.819 ms, which is equivalent to 146 graphs per second (GPS). 
In contrast, the worst results were achieved by the SplineConv model with a time of 92.762 ms (i.e., 10.78 GPS). 

\begin{table}[!t]
\caption{Operation computation times and number of graphs per second for each model.}
\resizebox{\columnwidth}{!}{%
\begin{tabular}{|l|c|c|}
\hline
Layer & Computation Time {[}ms{]} & Graphs per second \\ \hline
SplineConv & 92.762 & 10.78 \\ \hline
EdgeConv & 37.159 & 26.91 \\ \hline
GCNConv & 14.869 & 67.25 \\ \hline
SAGEConv & 6.819 & 146.65 \\ \hline
PointNetConv & 33.421 & 29.92 \\ \hline
\end{tabular}%
}
\label{times}
\end{table}

% \begin{table}[!t]
% \caption{Operation computation times and number of graphs per second for each model. The Spline convolution achieves the longest time, while the Edge and PointNet convolutions have values three times lower. The most efficient operation is the SAGE convolution.}
% \resizebox{\columnwidth}{!}{%
% \begin{tabular}{|l|c|c|}
% \hline
% Layer & Computation Time {[}ms{]} & Graphs per second \\ \hline
% SplineConv & 92.762 & 10.78 \\ \hline
% EdgeConv & 37.159 & 26.91 \\ \hline
% GCNConv & 14.869 & 67.25 \\ \hline
% SAGEConv & 6.819 & 146.65 \\ \hline
% PointNetConv & 33.421 & 29.92 \\ \hline
% \end{tabular}%
% }
% \label{times}
% \end{table}

\begin{figure}[!t]
\centerline{\includegraphics[scale=0.31]{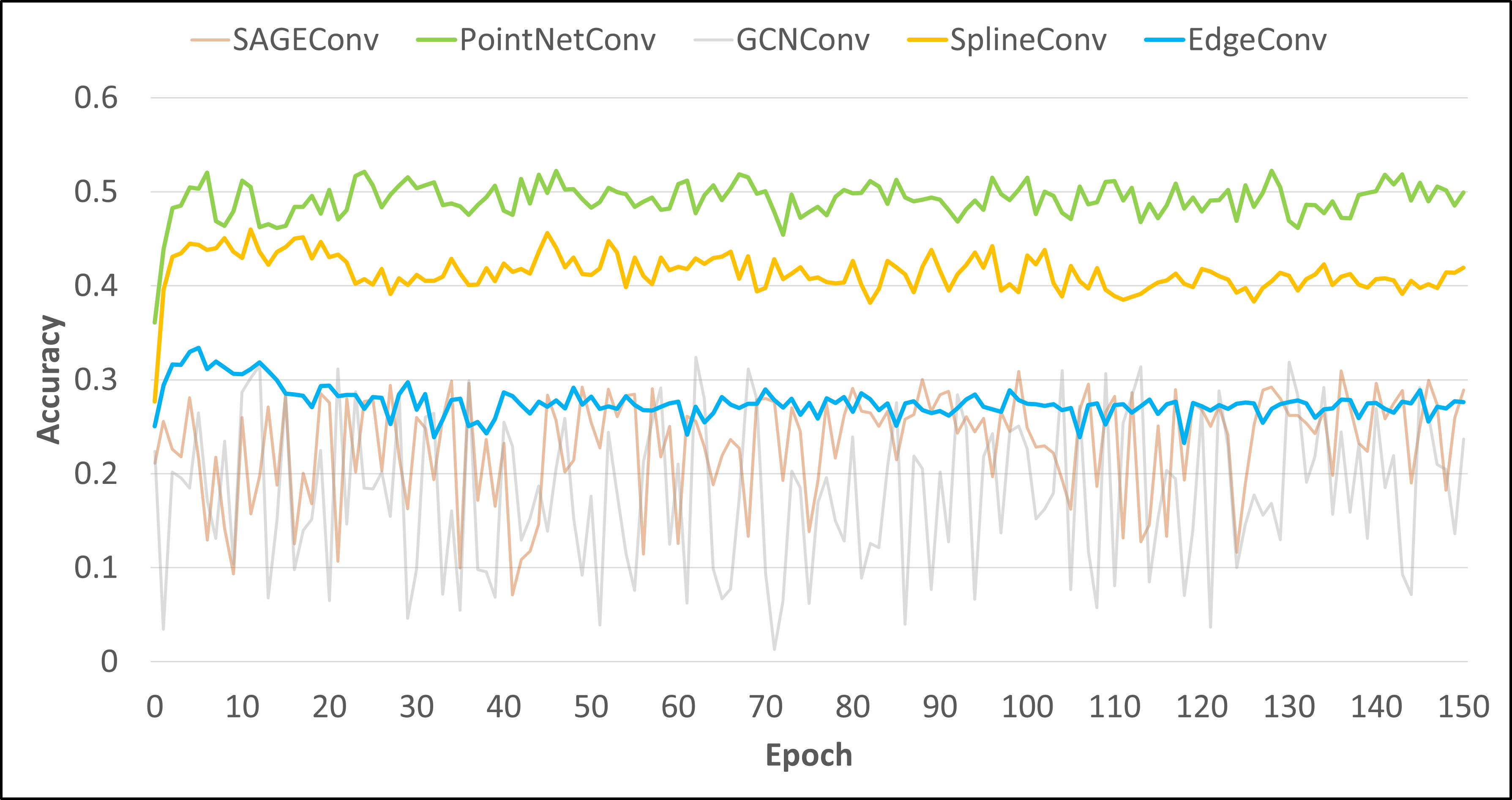}}
\caption{Accuracy for each operation. The PointNet convolution performed best, achieving a value of 52.3\%. The Edge Spline and Edge convolutions achieved 6.3\% and 18.9\% lower results, respectively. The GCN and SAGE training process did not lead to convergent results.} %TODO Komentarz
\label{acc_fig}
\end{figure}

The final test examined the results obtained by each model. 
For learning, the \textit{Adam} optimizer was used, with a \textit{learning\_rate} parameter of 1e-3 and a \textit{weight\_decay} value of 5e-3.  The number of epochs was limited to a maximum of 150, and the number of batches was 8. 
The learning results for each epoch are shown in Figure \ref{acc_fig}.
Surprisingly, the PointNetConv model had the best results, achieving a maximum accuracy of 52.3\%. 
The SplineConv model performed slightly worse, achieving an accuracy of 46\%. 
Meanwhile, the EdgeConv model, which does not use vertex position information, achieved the lowest accuracy of the three, at only 33.41\%. 
The learning process for the GCNConv and SAGEConv models did not lead to convergent results.

Based on all the results, it can be concluded that the PointNetConv model performed best, achieving the best results in the classification task and providing a good number of trainable parameters and a low execution time. 
On the other hand, the low results of the EdgeConv model show that information about the mutual position of vertices seems to be crucial in this problem.

%%%%%%%%%%%%%%%%%%%%%%%%%%%%%%%%%%%%%%%%%%%%%%%%%%%%%%%%%%%%%%%%%%%%%%%%%
\subsection{Detection model}\label{detection}

\begin{figure}[!t]
\centerline{\includegraphics[scale=0.19]{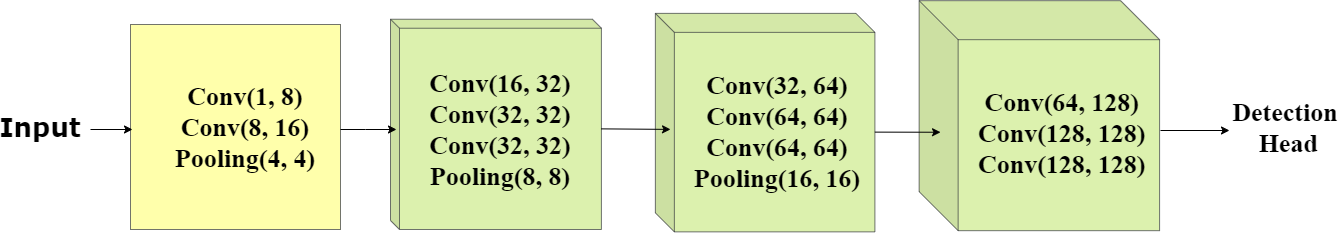}}
\caption{The proposed model created with PointNetConv blocks. The yellow block shows the module without residual connections. The green block has a residual connection between the 1st and 3rd convolution. The detection head consists of a flattening and a fully connected layers.}
\label{det_model}
\end{figure}

For the final analysis, we tested the PointNet operation in a detection task. We created our own model for feature extraction, which differs from the model used in the classification task. 
The architecture of this model is shown in Figure \ref{det_model}.

The model consists of two types of blocks. 
The first input block contains two convolutions, and MaxPooling with 4x4 window is used at the end. 
The other three blocks consist of three convolutions, and since the convolution operations do not affect the graph structure, a residual connection is used between the output of the first convolution and the output of the third convolution. 
MaxPooling is also used at the end of the first two blocks. In each subsequent block, the pooling window size is twice as large as the previous one.
In this way, we wanted to replicate traditional CNN models, which use many pooling layers with a 2x2 window size, and several convolution layers preceding them.

\begin{figure}[!t]
\centerline{\includegraphics[scale=0.33]{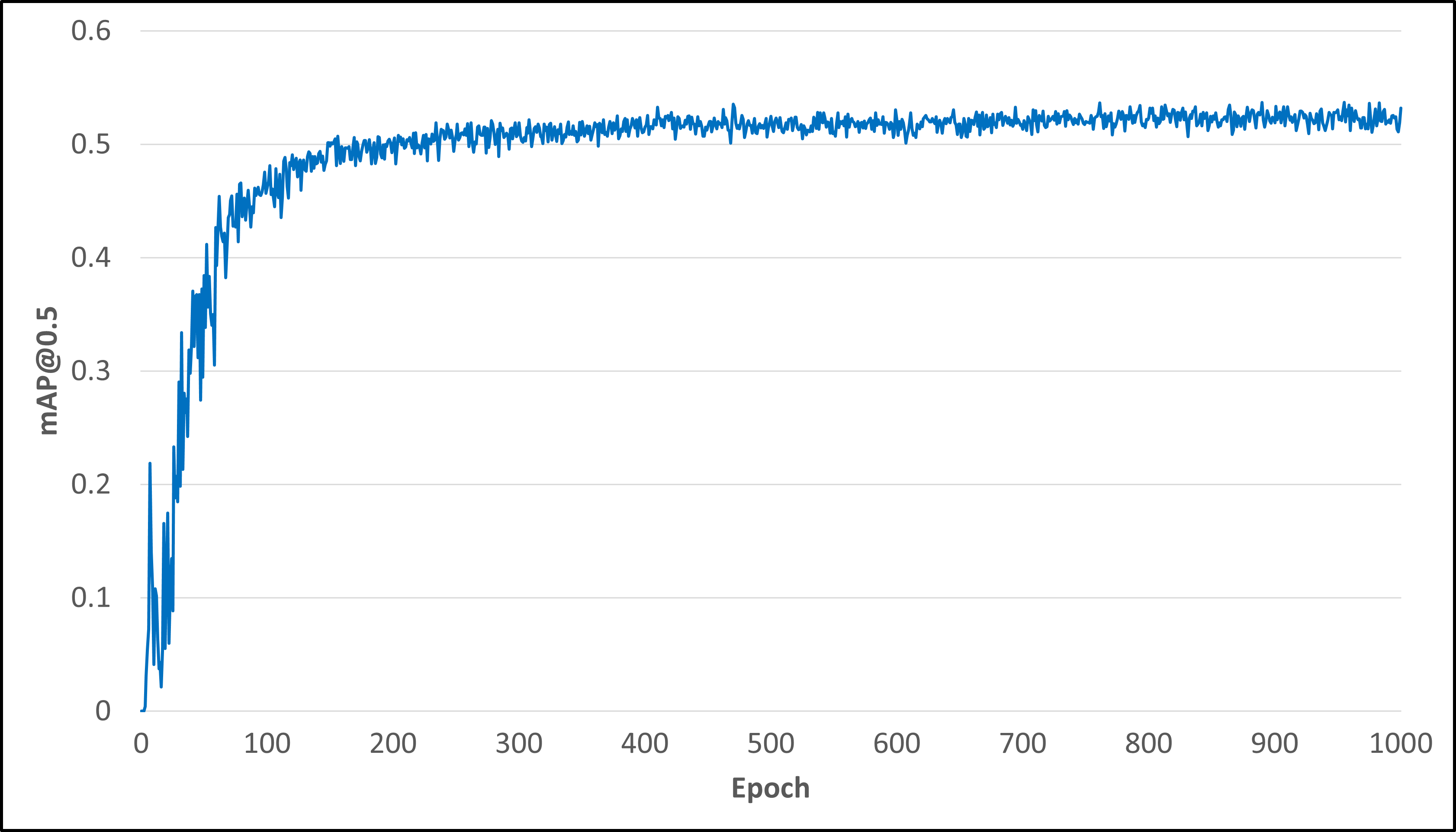}}
\caption{The graph shows the mean average precision (mAP) of the detection model at different epochs. Our proposed model achieved a score of 53.7\% on the N-Caltech101 dataset.}
\label{mAP}
\end{figure}

During the training process, we again used the \textit{Adam} optimizer with a \textit{learning\_rate} parameter of 1e-3 and a \textit{weight\_decay} factor of 1e-4. We set the maximum number of epochs to 1000, and set the batch size to 16. The learning results are shown in the Figure \ref{mAP}.

During the tests, we were able to achieve an accuracy of mAP@0.5 of 53.7 \%, as compared to the model in the \cite{b2} paper that achieves an accuracy of 59.5 \%, and the more complex architectures presented in the \cite{b3} paper achieve accuracies in the range of 62.9-73.2 \%.
This result was achieved using a model that has less than 100k trainable parameters in the feature extraction part. 
A time performance analysis performed on the entire validation set showed that this model achieves an average processing time of 12.186 milliseconds, which translates to an average of 82 graphs per second.

The tests performed allowed us to conclude that there is potential to improve current approaches in terms of memory consumption without significant loss in performance and satisfactory processing times.
The use of PointNet convolution brought not only a reduction in the number of model parameters, but also a reduction in memory requirements due to the fact that edge value information does not need to be stored. 

%%%%%%%%%%%%%%%%%%%%%%%%%%%%%%%%%%%%%%%%%%%%%%%%%%%%%%%%%%%%%%%%%%%%%%%%%%%%%%%%%
\section{Summary} \label{summary}

In our work, we presented a new perspective on event data processing using graph convolutional networks. In the experiments, we conducted an analysis of different convolution operators for graphs, taking into account the memory requirements and the number of trainable parameters of the models, while maintaining high performance and efficient operations. Our research has shown that the PointNet model allows a reduction in the number of trainable parameters by 450 times and a reduction in memory consumption for data representation by 4.5 times compared to state-of-the-art work. We also achieved relatively good classification results of 52.3\% accuracy and 53.7\% detection mAP@0.5 during network training, which are not significantly different from existing achievements.

Current research shows the great potential of graph convolutional networks in event data processing. However, our results show that focusing only on computational costs can lead to a significant increase in memory consumption. Thus, our work seeks to improve existing solutions and change the approach in future research, taking into account the important factor of memory usage.

In further research, we plan to continue to reduce memory consumption by modifying the graph description and using a more efficient data representation. The research will also include an analysis of other convolution operators and an attempt to develop our own operator, tailored specifically for event data. In addition, we intend to compare the benefits obtained for much larger data sets. Our plans also include the implementation of graph convolutional networks on hardware platforms, such as SoC FPGAs or Jetson Nano, in order to practically apply the model to real-world tasks.

%%%%%%%%%%%%%%%%%%%%%%%%%%%%%%%%%%%%%%%%%%%%%%%%%%%%%%%%%%%%%%%%%%%%%%%%%%%%%%%%%

\end{document}